\documentclass[11pt]{article}

\usepackage[final]{acl}

\usepackage{times,latexsym}
\usepackage{url}
\usepackage[T1]{fontenc}

\usepackage{etoolbox}
\AtBeginEnvironment{quote}{\par\singlespacing\small}

\usepackage{amsmath}
\usepackage{enumerate}
\usepackage{enumitem}
\usepackage{booktabs}
\usepackage{xcolor,colortbl,soul}
\usepackage{CJK}
\usepackage{adjustbox}
\usepackage{tabularx}
\usepackage[ruled,vlined]{algorithm2e}
\usepackage[font=small,skip=-5pt]{subcaption}
\usepackage{hyperref}

\usepackage[T1]{fontenc}

\usepackage[utf8]{inputenc}

\usepackage{microtype}

\usepackage{inconsolata}

\usepackage{graphicx}

\usepackage{xcolor}
\usepackage[many]{tcolorbox}    	%
\usepackage{setspace}               %
\usepackage{multicol} %
\usepackage{multirow} 
\usepackage{musicography}
\usepackage{tikz}
\usepackage{collcell}
\usepackage{xstring}
\usepackage{subcaption}
\usepackage{arydshln}

\definecolor{maria-color}{HTML}{7881F2}

\newcommand{\Espace}[1]{\mathcal{X}_{#1}}

\newcommand{\Topic}{t}

\newcommand{\Corpus}{\mathcal{D}_{\Model{}_{\Topic{}}}}
\newcommand{\Model}{m}

\newcommand{\Claims}{C}

\newcommand{\HSD}{D_{S}}
\newcommand{\divtype}{epistemic}
\newcommand{\Divtype}{Epistemic}

\newcommand{\codeurl}{\url{https://github.com/dwright37/llm-knowledge}}

\newtcolorbox{sharp_box}{
    sharpish corners, %
    boxrule = 0pt,
    toprule = 4.5pt, %
    enhanced,
    fuzzy shadow = {0pt}{-2pt}{-0.5pt}{0.5pt}{black!35} %
}

\newtcolorbox{prompt_box}{
    enhanced,
    boxrule = 0pt,
    colback = sub,
    borderline west = {1pt}{0pt}{main}, 
    borderline west = {0.75pt}{2pt}{main}, 
    borderline east = {1pt}{0pt}{main}, 
    borderline east = {0.75pt}{2pt}{main}
}

\title{What and Whose Knowledge?\\Measuring Epistemic Diversity in Large Language Models}

\author{
  Dustin Wright\textsuperscript{\musEighth{}} \hspace{0.3cm} Sarah Masud\textsuperscript{\musSharp{}} \hspace{0.3cm} Jared Moore\textsuperscript{\musFlat{}} \hspace{0.3cm} Srishti Yadav\textsuperscript{\musSharp{}} \hspace{0.3cm} Maria Antoniak\textsuperscript{\musNatural{}} \\
  \textbf{Peter Ebert Christensen\textsuperscript{\musSharp{}}} \hspace{0.5cm} \textbf{Chan Young Park\textsuperscript{\musDoubleSharp{}}} \hspace{0.5cm} \textbf{Isabelle Augenstein\textsuperscript{\musSharp{}}} \\
  \textsuperscript{\musEighth{}}Aalborg University Copenhagen \hspace{0.5cm}
  \textsuperscript{\musSharp{}}University of Copenhagen \hspace{0.5cm}  \textsuperscript{\musFlat{}}Stanford University\\
\textsuperscript{\musNatural{}}University of Colorado Boulder \hspace{0.5cm}
\textsuperscript{\musDoubleSharp{}}University of Texas Austin \\
\small{
    \textbf{Correspondence:} \href{dbw@cs.aau.dk}{dbw@cs.aau.dk}
    }
}

\begin{document}
\maketitle
\begin{abstract}
Large language models (LLMs) are increasingly used as primary knowledge sources, yet their epistemic diversity --- defined as the diversity of real-world claims in their outputs --- has never been measured. Low epistemic diversity would pose a risk of \textit{knowledge collapse}
as homogeneous LLMs mediate a shrinking in the
range of accessible information over time. The dominant paradigm is that overall LLM diversity is low, but this is always with respect to a single point in time, with no reference baseline or consideration for variation across countries. We address this gap in knowledge by performing the first systematic study of epistemic diversity in LLMs across time and cultural context, testing 27 LLMs on 155 topics covering 12 countries, resulting in 1.7M responses and 70M individual claims. We find that epistemic diversity has increased substantially over the past three years, a positive counter to recent diversity pessimism. However, despite progress, we find that every system is less diverse than a search baseline. This gap is not uniform: RAG can improve diversity, while large models are counterintuitively less diverse than smaller ones. Moreover, LLM parametric knowledge systematically reflects English over local-language knowledge for country specific topics. Together, these results demonstrate that while progress on epistemic diversity is tangible, it is insufficient and unevenly distributed.\footnote{Code and data: \codeurl{}}

\end{abstract}

\section{Introduction}
\begin{figure}[th]
        \centering
        \includegraphics[width=0.98\linewidth]{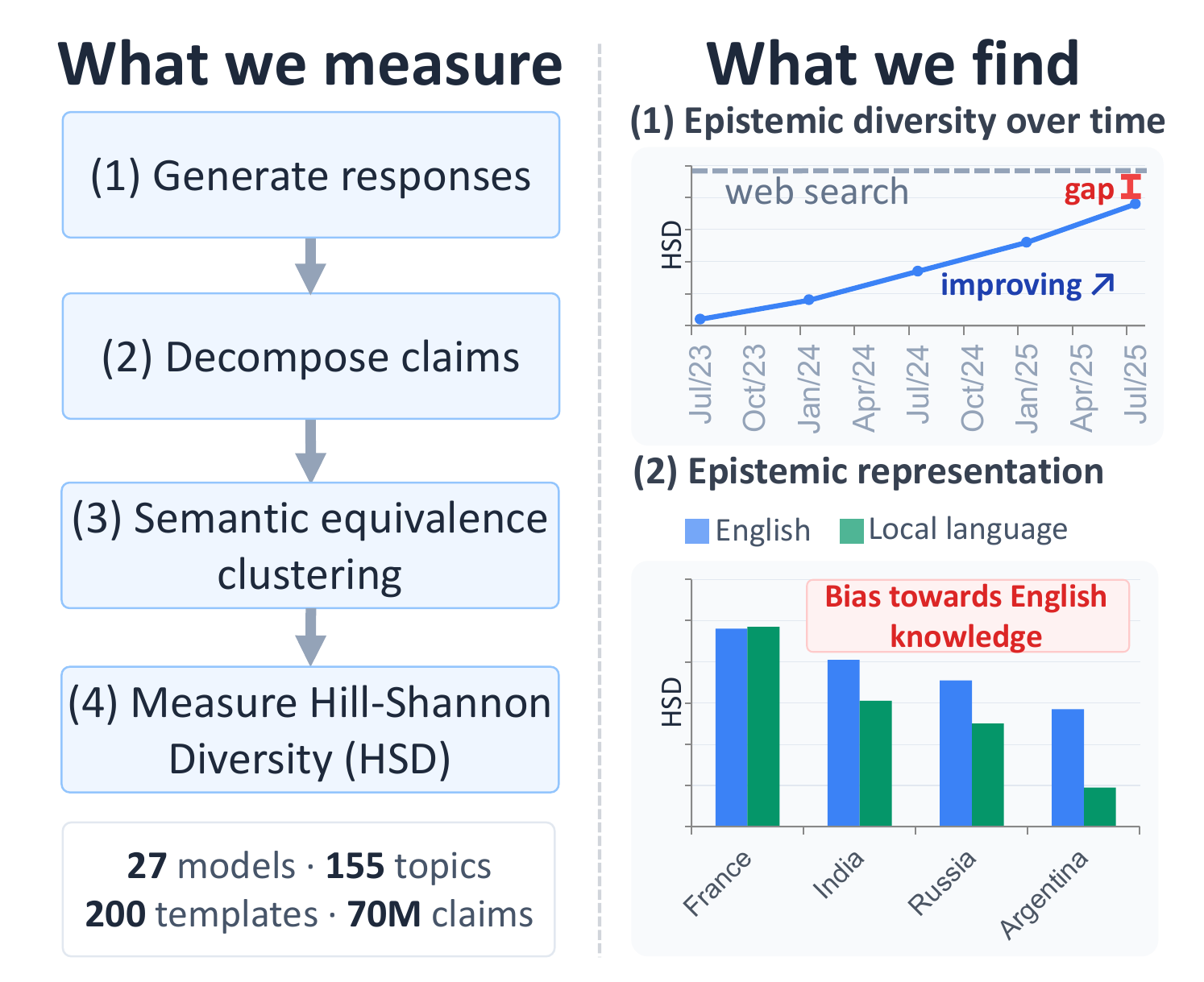}
        \caption{In this work, we measure \divtype{} diversity via variability in claims about the world. We compare across models, settings, and country-specific topics.
        }
        \label{fig:fig1}
\end{figure}

Large language models (LLMs) are widely used for knowledge-intensive tasks such as writing assistance~\cite{sun2025large}, summarization~\cite{DBLP:journals/corr/abs-2502-14409}, and research~\cite{DBLP:conf/iclr/Si}. 
Search interfaces are now prioritizing ``AI Overviews'' to answer queries.
And there has been speculation that people will soon access most information through an LLM intermediary~\cite{DBLP:journals/ais/Peterson25}.

At the same time, LLM outputs are homogeneous along multiple axes~\cite{sourati2025homogenizing}. They reflect only a narrow range of writing and reasoning styles, use a limited vocabulary~\cite{DBLP:journals/corr/abs-2502-11266}, and convey only certain semantics~\cite{DBLP:journals/corr/abs-2503-20062,DBLP:journals/corr/abs-2510-22954,DBLP:conf/nips/YuZZMRKSZ23}. Given the tasks people use these models for, there is a risk of knowledge collapse --- where ``access to the original diversity of human knowledge is increasingly mediated by a partial and increasingly narrow subset of views''~\cite{DBLP:journals/ais/Peterson25}.
There is thus an urgent need for reliable measurements of \textbf{\divtype{} diversity} of LLMs, which we define as the diversity of claims about the world in a set of LLM outputs.
Moreover, as the dominant paradigm posits static or worsening diversity in LLM outputs~\cite{DBLP:journals/corr/abs-2510-22954}, there is a need to study diversity patterns in LLMs across models, time periods, and cultural context to gain insight into the current and future risks of knowledge collapse and how to prevent it.

To this end, we perform a first empirical study measuring the \divtype{} diversity in LLM outputs (see \autoref{fig:fig1}).
We investigate diverse scenarios in which LLMs are prompted in different ways about the same topics, generating open-ended free-text responses with multiple embedded claims. Concretely, we (1) sample responses from 27 LLMs using multiple versions, settings, and topics with human-written prompt templates collected from \citet{DBLP:journals/corr/abs-2502-08395}, (2) decompose individual LLM responses into one or more claims, (3) partition those claims into semantically equivalent clusters~\cite{DBLP:journals/nature/FarquharKKG24}, and (4) quantify the diversity of the sample of claims with Hill-Shannon diversity, a widely used metric for sample diversity measurement in ecology~\cite{roswell2021conceptual}. 
We measure diversity across and within 155 general domain and country specific topics for 12 countries, examining both \textit{what} claims are generated and \textit{whose} knowledge is represented by the generated claims in terms of country. In doing so, we produce a dataset of 70M claims clustered by semantic equivalence and generated by a set of the most popular LLMs from recent years.

Encouragingly, we find that 
\divtype{} diversity in LLM outputs is generally \textit{increasing} over time, a positive counter to recent work positing knowledge collapse~\cite{DBLP:journals/ais/Peterson25} and static lack of diversity in LLMs~\cite{DBLP:journals/corr/abs-2510-22954}. However, despite this progress, we find that there is a gap in epistemic diversity between all systems in our study and a weak traditional search baseline. This diversity gap between web search and LLMs is particularly alarming given recent shifts in web search toward AI-exclusive designs.\footnote{\href{https://techcrunch.com/2026/05/19/google-search-as-you-know-it-is-over/}{https://techcrunch.com/2026/05/19/google-search-as-you-know-it-is-over/}} 

Further, this gap is not uniform. Retrieval-augmented generation (RAG) has a statistically significant \textit{positive} impact, highlighting the importance of RAG in preventing knowledge collapse. However, similar to contemporary work~\cite{zhang2025noveltybench,DBLP:journals/corr/abs-2410-22592}, we counterintuitively find that model size has a statistically significant \textit{negative} impact --- smaller models generate more diverse knowledge than larger ones.
Finally, when considering the countries that topics are associated with, we find that some countries may experience greater risks of knowledge collapse than others.
RAG has an uneven effect: certain countries such as the USA see more benefit due to a greater diversity in their RAG sources, which are not available for many languages. And compared to Wikipedia in both English and local languages for country-specific topics, we find that LLM parametric knowledge reflects English more than local language knowledge, highlighting a gap in whose knowledge is represented by LLMs in terms of country. Overall, our results demonstrate that while progress on epistemic diversity is tangible, it is insufficient and unevenly distributed.

\section{Background}
LLMs are increasingly being used for knowledge-centric tasks~\cite{DBLP:journals/tkdd/YangJTHFJZYH24}, and their outputs can influence people's behavior~\cite{DBLP:conf/candc/AndersonSK24, DBLP:conf/chi/JakeschBBZN23,bai2025llm}. 
If we become reliant on AI systems for these tasks, a lack of diversity in AI systems' outputs could also reduce the diversity in our collective knowledge. 
Our work studies this risk through the lenses of LLM homogenization and knowledge collapse.

\paragraph{LLM Homogenization}
A robust body of work shows that LLMs suffer from a lack of diversity along many dimensions \cite{sourati2025homogenizing,DBLP:journals/tacl/GuoSC25,DBLP:conf/acl/IveyKLSRRZAKYWI26}. These include: lexical and stylistic~\cite{DBLP:journals/corr/abs-2502-11266,DBLP:conf/emnlp/ShaibELW24, DBLP:conf/iclr/Padmakumar024}, semantic~\cite{DBLP:journals/corr/abs-2503-20062,DBLP:journals/corr/abs-2510-22954,DBLP:conf/nips/YuZZMRKSZ23, DBLP:conf/iclr/Padmakumar024,moon2025homogenizing}, creative~\cite{DBLP:journals/corr/abs-2501-00273,moon2025homogenizing,DBLP:journals/corr/abs-2501-19361}, conceptual~\cite{DBLP:conf/naacl/MurthyUH25}, recommendation~\cite{DBLP:conf/iui/DudyTRALB25,DBLP:journals/corr/abs-2404-15149,DBLP:journals/corr/abs-2506-00074,ShurOfry2024GrowingATA}, coding~\cite{shypula2025evaluating}, and perspective diversity~\cite{DBLP:conf/emnlp/0001ABYBA24,abdurahman2024perils,zhang2025noveltybench,DBLP:journals/corr/abs-2306-16388,DBLP:conf/acl/RottgerHPHKSH24,DBLP:conf/emnlp/MooreDY24,DBLP:conf/emnlp/HayatiLRK24,DBLP:conf/emnlp/WangPLB25}, among others. Our work is most similar to perspective diversity, which has shown that LLM-generated views, opinions, and beliefs tend to reflect only a small subset of the world~\cite{DBLP:journals/corr/abs-2306-16388,atari2023humans,abdurahman2024perils,DBLP:journals/jbd/AlveroLRKJA24}. 
We build on this by looking at the diversity of all semantically equivalent classes of \textit{claims} in open-ended free-text LLM responses, avoiding shallow and/or fuzzy features used in semantic and perspective similarity. Our methodology and empirical results show how this diversity has changed over time and across countries, with actionable recommendations, and offering more general conclusions about knowledge diversity than previous work.

\paragraph{Knowledge Collapse}
A growing concern among scholars is that LLM homogenization, combined with increased adoption of LLMs, will lead to epistemic problems at a societal level~\cite{zheng2023epistemological,messeri2024artificial,DBLP:journals/ais/Peterson25,wagner2025death,DBLP:journals/corr/abs-2506-06166,farrell2025large}. 
\citet{DBLP:journals/ais/Peterson25} defines ``knowledge collapse'' as LLMs facilitating a dwindling of knowledge into an increasingly narrow set of ideas.
Knowledge collapse may affect existing knowledge sources such as Wikipedia~\cite{wagner2025death}, erase minoritized knowledge~\cite{zheng2023epistemological}, pollute scientific discoveries~\cite{messeri2024artificial}, hamper political discourse~\cite{coeckelbergh2025llms}, and limit ideation in writing~\cite{DBLP:conf/candc/AndersonSK24}. These concerns echo other contemporary epistemic issues such as ``popularity bias''~\cite{ciampaglia2018algorithmic}, and ``filter bubbles''~\cite{DBLP:conf/www/NguyenHHTK14} occurring with the use of recommender systems.
Our work assesses this risk through a new methodology for measuring \divtype{} diversity and an empirical study of this phenomenon in LLMs. Using this, we compare risks in LLM outputs across models, sources, and time, and we contrast our findings with measurements for traditional knowledge systems.

\section{Problem Setup and Notation}
\label{sec:motivation_notation}
We measure the \divtype{} diversity of LLMs as the diversity of the distribution of claims made in their outputs to open-ended prompts on various topics.
Concretely, we develop a new methodology which decomposes open-ended LLM-responses into claims, partitions these claims into semantically equivalent clusters, and measures the Hill-Shannon diversity of these clusters, a statistically grounded measure of diversity used widely in ecology for measuring species diversity~\cite{DBLP:journals/ais/Peterson25,roswell2021conceptual}.

To describe our approach (\autoref{fig:fig1}), we adopt the following notation. First, a corpus of text $\Corpus{}$ is elicited about a topic $\Topic$ from a model $\Model{}$. 
$\Corpus{}$ contains free text which can be decomposed into a list of $n$ claims $\Claims_{\Model{}_{\Topic}}$.
These claims can be organized into a set of unique \textit{meaning classes} $\Espace{\Model{}_{\Topic}}$. A unique meaning class contains claims that
are semantically equivalent to one another but not to claims in other classes. Then, $x_{i}$ is defined as the empirical frequency of meaning class $i$ enumerated from $\Claims_{\Model{}_{\Topic}}$, $p_{i}$ is the probability of meaning class $i$ calculated as the relative frequency of $i$ in the sample (i.e., $\frac{x_{i}}{n}$), and $P_{\Model{}_{\Topic}}$ is a categorical distribution with $p_{i} \in P_{\Model{}_{\Topic}}$ for all meaning classes $i$.

\section{Data Collection}
\label{sec:data}

We first describe how $P_{\Model{}_{\Topic}}$ is constructed for a particular model $\Model{}$ and topic $\Topic{}$ (for details about the specific models and topics we study see \autoref{sec:empirical_results}). Acquiring $P_{\Model{}_{\Topic}}$ involves a three step process similar to \citet{DBLP:conf/emnlp/0001ABYBA24} and \citet{zhang2025noveltybench}:

\begin{enumerate}[noitemsep, leftmargin=*]
    \item \textbf{Generate}: Acquire a corpus $\Corpus{}$ of free-text LLM responses to natural input prompts.
    \item \textbf{Decompose}: Decompose $\Corpus{}$ into a list of claims $\Claims_{\Model{}_{\Topic}}$.
    \item \textbf{Cluster}: Group the list of claims into meaning classes $\Espace{\Model{}_{\Topic}}$ whose members are semantically equivalent, then calculate $P_{\Model{}_{\Topic}}$.
\end{enumerate}

\subsection{Generation and Decomposition} 
\label{subsec:gen_decompose}

To elicit open-ended responses about any given topic $t$, we start with the prompt templates curated in \citet{DBLP:journals/corr/abs-2502-08395}. 
These templates take forms such as ``write me an essay about \texttt{\{topic\}},'' where \texttt{\{topic\}} is replaced by $t$. From the original 1,000 templates, we select a subset of 479 that focus on information seeking by manually filtering out NSFW (not safe for work) templates, creative writing templates such as ``write a 50's soviet style song about $\Topic$,'' outlines such as ``write an index for a book on $\Topic$,'' or those explicitly asking for a list of references.
The filtering is performed by three of the paper's authors based on group consensus about the appropriateness of each prompt template for eliciting knowledge claims.
We then randomly sample 200 templates, where every template is used for every topic in the study.

After generating responses we \textbf{decompose} them into individual claims $\Claims_{\Model{}_{\Topic}}$ using Llama 3.1 70B, 
prompting the model to decompose the response to the individual claims which are explicitly relevant for the topic,  using non-overlapping chunks of three sentences at a time. We thus ensure that each input is self-contained with all necessary context while capturing all claims that are explicitly about the given topic. 

We initially developed three decomposition prompts (P1-P3; see \autoref{sec:prompts} for prompt text) and selected the best one based on an LLM-as-a-judge setup~\cite{DBLP:conf/emnlp/LiuIXWXZ23}. Two human annotators labeled 100 data points for each of two tasks: (1) rating the \textit{quality} of individual decomposed claims on a Likert scale from 1-5 and (2) quantifying how many relevant claims in the original input chunk are \textit{missing} from the list of decomposed claims. 

We achieved moderate to strong Pearson $r$ across our two annotators (0.74, 0.58 respectively), and a third annotator broke annotation ties. Our prompts for LLM-as-a-judge (see \autoref{sec:annotation_instructions} for details) achieved Pearson correlations of 0.60 and 0.68 respectively.
We use this setup to automatically label a sample of 6K instances across the three decomposition prompt variants. The results of this analysis are given in \autoref{tab:decomposition_eval}. We use P3 for the final decomposition as claim fidelity is more important than missing claims in our study: we eventually downsample the set of claims to match the coverage of each distribution (see \S\ref{sec:diversity} for details).

\begin{table}[t]%
    \centering
    \rowcolors{2}{gray!15}{white}
    \begin{tabular}{r c c c}
    \toprule %
    & \textbf{P1} & \textbf{P2} & \textbf{P3} \\
    \midrule
    
    \cellcolor{black!75}\color{white}Quality$\uparrow$ & $4.58$ & $4.67$ & $\mathbf{4.69}$ \\
    \cellcolor{black!75}\color{white}Missing Claims$\downarrow$ & $\mathbf{0.06}$ & $0.24$ & $0.25$  \\

    \bottomrule %

    \end{tabular}
    \caption{Performance of different prompts for claim decomposition using LLM-as-a-judge (\autoref{subsec:gen_decompose}).}
    \label{tab:decomposition_eval}
\end{table}

\subsection{Clustering}
\label{sec:clustering}

Following \citet{DBLP:journals/nature/FarquharKKG24}, we create clusters based on mutual entailment using a strong pretrained model for natural language inference.\footnote{HuggingFace ID: microsoft/deberta-large-mnli} %
This is done in a greedy bootstrap fashion, building clusters one claim at a time by assigning a given claim to an existing cluster if it mutually entails at least one other claim in that cluster. To reduce the overhead of measuring mutual entailment across all $2n^{2}$ pairs for a set of $n$ claims, we check entailment between each claim and the $N$ most similar claims to it (i.e., $2N*n$ comparisons) measured using an S-BERT model.\footnote{HuggingFace ID: all-MiniLM-L6-v2} We then perform a post-processing step to break up larger clusters using DBSCAN~\cite{DBLP:conf/kdd/EsterKSX96}, a related clustering approach~\cite{DBLP:journals/ais/Peterson25, DBLP:conf/emnlp/0001ABYBA24}. See \autoref{sec:algorithm} for the algorithm pseudocode.

We evaluate clustering using LLM-as-a-judge, again (1) rating cluster quality (Likert score 1-5) and (2) predicting potential missing claims (i.e., unclustered claims which should have been included in a given cluster). 
Two annotators labeled data for both tasks, achieving moderate agreement (0.44 Pearson $r$ for cluster quality on 100 examples; 87\% agreement accuracy for missing claims on 360 examples), followed by a third annotator to break ties. We develop prompts that achieve 0.68 Pearson $r$ with the human labels for cluster quality and 83.5 weighted F1 score for predicting if a claim should have been clustered. See \autoref{sec:annotation_instructions} for full evaluation setup. 15K samples of 3-5 sentences are drawn from clusters acquired using our clustering algorithm and automatically labeled, where semantic similarity for task (2) is measured using S-BERT. We acquire a cluster quality score of 4.17 and a missing rate of 13.79\%, indicating that the cluster quality is generally strong and misses only a relatively small proportion of unclustered claims. See \autoref{sec:sem_sim_compare} for a comparison to using semantic-similarity only.

\section{Measuring \Divtype{} Diversity}
\label{sec:diversity}

After obtaining the set of unique meaning classes $\Espace{\Model{}_{\Topic}}$ and their probabilities $p_{i} \in P_{\Model{}_{\Topic}}$ for a model $\Model{}$ on topic $\Topic$, we must quantify its diversity to rank different models. There are a number of possible choices to measure diversity based on categorical distributions~\cite{DBLP:journals/corr/abs-2507-20858}, for example, richness (the number of unique meaning classes) or entropy. However, we desire a measure which accounts for the distribution of claims (ignored by richness) and allows for meaningful comparison between settings (impossible for entropy due to its logarithmic scaling). An appropriate measure of diversity that accounts for both of these issues is Hill-Shannon diversity~\cite{hill1973diversity,jost2006entropy}, a special form of the general Hill diversity which balances emphasis between frequent and rare claims. It additionally satisfies the \textit{replication principle}, where relative changes in categories and abundances between samples result in proportionate changes to the score~\cite{roswell2021conceptual}. It is calculated as the exponentiated entropy of the categorical distribution $P_{\Model{}_{\Topic}}$: 
\begin{equation}
\label{eq:hsd}
    \HSD{}(P_{\Model_{\Topic}}) = \exp\{-\sum_{i}^{{|P_{\Model_{\Topic}}}|}p_{i}\ln p_{i}\}
\end{equation}

This measure has an intuitive interpretation~\cite{roswell2021conceptual}: from \autoref{eq:hsd} we can see that, since $e^{x}$ is a monotonically increasing function of $x$, HSD is maximized when the entropy is maximized, i.e., under a uniform distribution.
Consider a distribution $P_{\Model_{\Topic}}$ generated by a model $m$ for topic $t$ with $|P_{\Model_{\Topic}}| = K$ (i.e., $K$ meaning classes). Under a uniform distribution, the HSD resolves to: 
\begin{align*}
     \HSD{}(P_{\Model_{\Topic}})
     &= \exp\{-\sum_{i}^{K}\frac{1}{K}\ln \frac{1}{K}\} \\ 
     &= \exp\{\ln K\} = K
\end{align*}
This means that if $|P_{\Model_{\Topic}}| = K$ then $\HSD{}(P_{\Model_{\Topic}}) = \hat{K} \leq K$. When $\hat{K}$ is strictly less than $K$, we can then view the effective diversity of that distribution of claims as equivalent to a distribution $P_{\hat{m}_{\Topic}}$ which has a uniform probability of generating one of $\hat{K}$ meaning classes. This implies that one can, for example, claim a model which generates a claim distribution with an HSD of $2\hat{K}$ is effectively twice as diverse as a model with an HSD of $\hat{K}$.

The way one samples the claims, $C_{\Model{}_{\Topic}}$, has a large impact on $\HSD{}$. This can occur despite ``equal-effort'' sampling, e.g., prompting all models with the same number of prompts, because the long tail of claims may vastly differ between settings. To illustrate: consider two models, one with low diversity ($a$) and one with high diversity ($b$), from which we sample the same number of claims, $n$; $a$ is less diverse, so $n$ samples may be sufficient to fully characterize $P_{a_{\Topic}}$, and sampling more claims is unlikely to have a substantial impact on $\HSD{}(P_{a_{\Topic}})$. On the other hand, sampling more claims from $b$ may introduce new meaning classes to $P_{b_{\Topic}}$ which take longer to uncover due to their rarity, thus increasing $\HSD{}(P_{b_{\Topic}})$. Therefore, by sampling $\approx n$ claims for both models, we have potentially overestimated the diversity of $a$ relative to $b$.

To account for this, we use a combination of coverage estimation and rarefaction~\cite{chao2012coverage,roswell2021conceptual}. Coverage describes how completely we have covered the true distribution and is calculated based on the frequency of meaning classes in $\Espace{\Model{}_\Topic{}}$ using the estimator from \citet{chao2012coverage}:
\begin{equation}
\label{eq:coverage}
    V(\Espace{\Model{}_\Topic{}}) = 1 - \frac{f_1}{n}\left[\frac{(n-1)f_{1}}{(n-1){f_1} + 2f_{2}} \right],
\end{equation}
where $n$ is the number of claims, $f_{1}$ is the number of singleton classes in $\Espace{\Model{}_\Topic{}}$ (i.e., the number of meaning classes $i$ where $x_{i} = 1$), and $f_{2}$ is the number of doubleton classes (i.e., the number of meaning classes $i$ where $x_{i} = 2$). When we wish to compare the diversity of models on a topic $t$, we use \autoref{eq:coverage} to calculate the coverage of all models and rarefy each set of claims (i.e., downsample) to the minimum coverage achieved across models.

\begin{figure*}[t]
        \centering
        \includegraphics[width=0.98\linewidth]{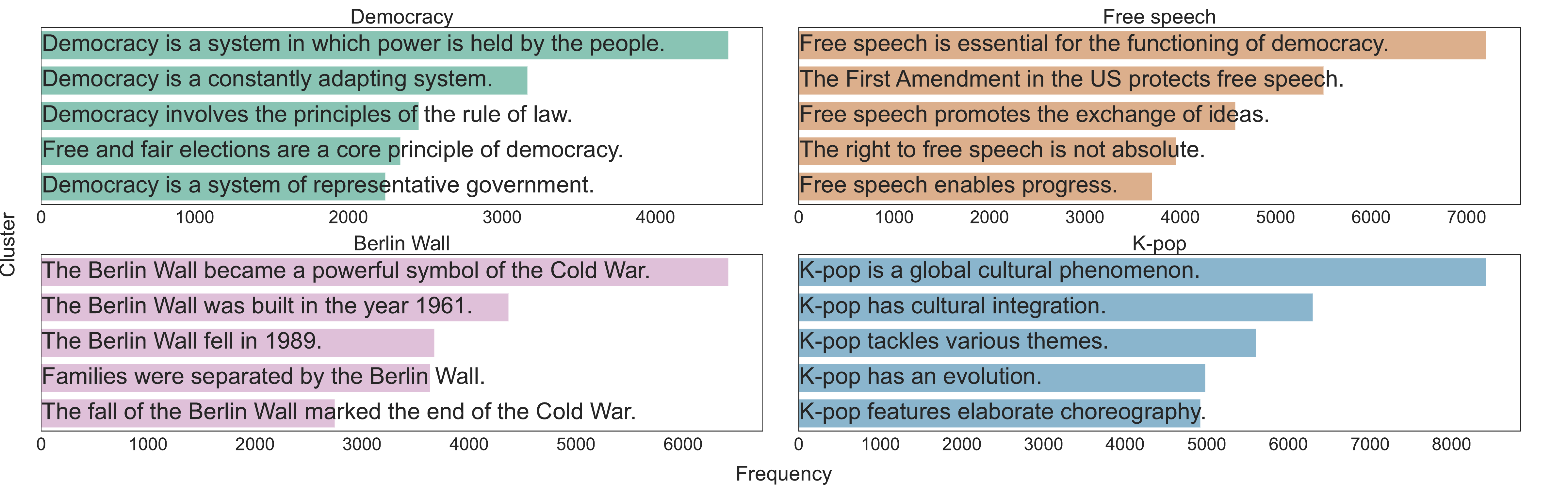}

        \caption{Histograms of the top five clusters for four topics after generating text, decomposing, and clustering decomposed claims across all models in our study. The frequency of claims in each cluster, $x_{i}$, is represented by the colored bars. By the 5th cluster, $x_{i}$ is halved for all four topics, indicating a large decay rate for $x_{i}$. The top clusters convey broad and general information for each topic.}
        \label{fig:top_clusters}
\end{figure*}

\begin{figure*}[th]
        \centering
        \includegraphics[width=0.98\linewidth]{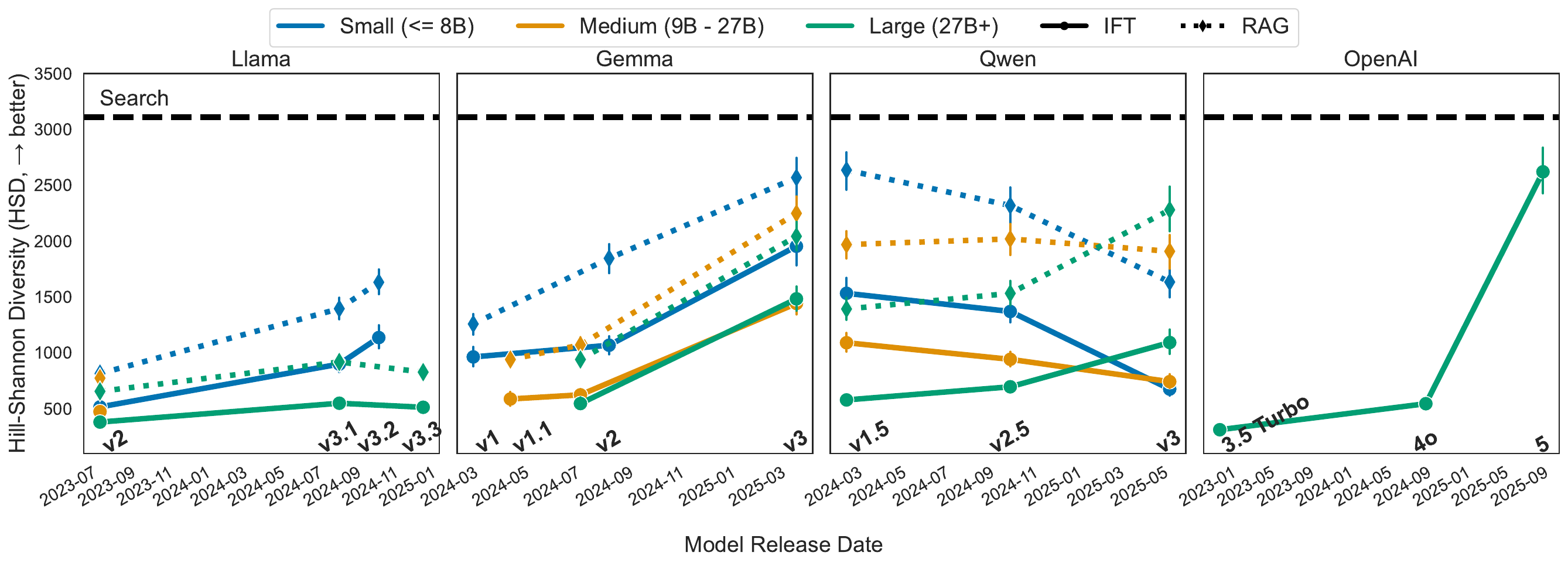}
        \caption{\Divtype{} diversity
       vs. model release date. Each point is a single model, with lines connecting models of approximately the same size across released versions. Error bars are 95\% bootstrapped confidence intervals based on the HSD of each topic (N=155). Absolute diversity is low for all models compared to a very modest search baseline. However, for all families except Qwen and most sizes, we see a trend of improved diversity.
        }
        \label{fig:id_by_model}
\end{figure*}

\section{Empirical Results}
\label{sec:empirical_results}
How epistemically diverse are LLMs? To answer this broad question, we pose several sub-questions focused on how \divtype{} diversity has changed over time \textbf{(RQ1)}, how generation setting impacts \divtype{} diversity \textbf{(RQ2)}, how size and model family impact \divtype{} diversity \textbf{(RQ3)}, and how the topic associated with a country impacts diversity and representation \textbf{(RQ4)}. %

\subsection{Models, Settings, and Topics}
\label{sec:topics_models}

\paragraph{Models and Settings} We select 27 systems (open-source LLMs and closed-source systems) across 4 families: Llama (versions 2, 3.1, 3.2, and 3.3), Gemma (versions 1, 1.1, 2, and 3), Qwen (versions 1.5, 2.5, and 3), and OpenAI (GPT 3.5 Turbo, 4o, and 5), with small, medium, and large model variants, thus including multiple release dates (between late 2022 and 2025) and sizes (1B to 72B). \autoref{tab:model_card} provides the details of all models.
We test each model with two different settings: (1) parametric memory with instruction-fine-tuning only (IFT) and (2) retrieval augmented generation (RAG). In addition, we include as a baseline the top 40 web pages retrieved from Google search for each topic.\footnote{Search performed on January 1, 2026 in the US region. 
} 

For RAG we use the top 20 search results per topic as our database from which we select top paragraphs, reflecting a common setup with precedent in the literature~\cite{DBLP:journals/corr/abs-2203-05115,DBLP:conf/acl/VuI0CWWTSZLL24}. For retrieval, we match prompts to the top paragraphs using S-BERT embeddings, including up to 1000 tokens of context from these retrieved paragraphs. We additionally shuffle the ranks of all paragraphs with a cosine similarity to the query above the mean cosine similarity across all paragraphs and prompts in the data to observe how varying context impacts output diversity in the RAG setting. For generation, we use top-$p$ sampling (top-$p = 0.9$, temperature $= 1.0$) and generate a maximum of 2,100 tokens per prompt ($\approx$ the median number of tokens in our web pages).

For rarefaction, we calculate the coverage of each sample for each setting on each topic using \autoref{eq:coverage}, and rarefy all samples to the minimum.
The search result baseline has much lower coverage than the systems in our study, so we only rarefy search if the coverage is above the minimum achieved by any system. As such, our search results are a lower-bound baseline; the true diversity of search relative to the systems tested is higher.

\paragraph{Topics} 
We use a sample of 30 general topics from IssueBench~\cite{DBLP:journals/corr/abs-2502-08395} and 125 additional hand-curated topics.
These include 10-13 country-specific topics across 12 countries about important figures and historical events to help measure the impact of country-specific context.
To ensure that these are well documented and known topics, we only select topics where their associated English Wikipedia page has a content rating of at least ``C''.
\footnote{\url{https://en.wikipedia.org/wiki/Wikipedia:Content_assessment}} 
Following IssueBench, these include a mix of controversial and non-controversial topics, e.g., general topics for which there may be contentious views, like nuclear weapons and pornography, and events which may be subject to varied interpretations. All events occurred before the earliest knowledge cutoff of any model in our study. In total, we have 155 topics (see \autoref{sec:topics} for the list), yielding 1.7M responses prior to decomposition and 70M claims after decomposition.

\paragraph{Clustering} 
For our clustering algorithm we have one parameter to select: the number of claims $N$ to retrieve for comparison. To choose this, we first ran a sample of 50,000 claims from 10 topics (5,000 each) through \autoref{alg:clustering}, setting $N=10$. From the list of top 10 most similar claims, we measure mutual entailment with the retrieving claim, and count the number of times that the $k$\textsuperscript{th} most similar claim ($k \in [1,10]$) is mutually entailed. We found that when a claim is matched, $98.4$\% of the time it is within the first 6 most similar claims, and therefore select $N=6$ for our experiments.
We give examples of the top 5 clusters for four topics (democracy, free speech, Berlin Wall, and K-pop) in \autoref{fig:top_clusters}. 
The frequency of claims drops rapidly; 
high-frequency clusters unsurprisingly provide broad, common information about each topic.

\subsection{\Divtype{} Diversity Across Time}

First, we focus on \textbf{RQ1}: \textbf{How has \divtype{} diversity in LLMs changed over time?} We measure \divtype{} diversity across all topics and settings mentioned previously, and plot topic-averaged Hill-Shannon diversity (HSD) vs. model release date in \autoref{fig:id_by_model}.
We plot traditional search diversity as a baseline and separately plot IFT and RAG to see the impact of generation setting. We connect models over time based on their relative size and version.

Our empirical results reveal that, with the exception of Qwen, \divtype{} diversity is increasing over time, countering prevailing narratives around static lack of diversity~\cite{DBLP:journals/corr/abs-2510-22954} and knowledge collapse~\cite{DBLP:journals/ais/Peterson25}. This is particularly pronounced for recent models: Gemma 3 and GPT-5 show the largest single-generation gains in our study, suggesting that recent model developments are positively contributing to epistemic diversity.

\begin{table}[t]
\centering
\setlength{\tabcolsep}{2pt}
\rowcolors{2}{gray!15}{white}
\begin{tabular}{@{}lll@{}}
\toprule
Predictor  & $\widehat{\beta}$ & $p$-value         \\ \midrule
\cellcolor{black!75}\color{white}Intercept (IFT)  &  939.90($\pm223.67$) & $p\ll1\mathrm{e}{-3}$ \\ \midrule
\cellcolor{black!75}\color{white}RAG   & +656.50($\pm33.83$)& $p\ll1\mathrm{e}{-3}$ \\
\cellcolor{black!75}\color{white}Search & +2170.85($\pm1183.57$)& $p\ll1\mathrm{e}{-3}$          \\
\bottomrule
\end{tabular}

\caption{Estimated coefficients ($\hat{\beta}$) w/ 95\% CIs, and $p$-values of a linear mixed effects model with the setting (IFT, RAG, Search) as fixed effects and the model as random effects. The dependent variable is HSD. RAG and Search are statistically significantly more diverse than IFT.}
\label{tab:ift_vs_rag}
\end{table}

Despite this, every model in our study generates less diverse claims than our search baseline.
This means that people in aggregate will see more diverse information about our topics by reading the top web pages rather than prompting an LLM in different ways.
The search baseline is also a \textit{weak} baseline, as we use only up to the top 40 Google search results for each topic based on simply using the topic as a search term (e.g., ``democracy''). 
Further, as mentioned in
\autoref{sec:topics_models}, 
because we rarefy to minimum coverage and only rarefy search when it exceeds that minimum, the reported gap is a lower bound. In absolute terms, we can thus conclude that \textit{search is at least 18.7\% more diverse than our best model (GPT-5) in terms of the claims it generates} (average HSD of 3110 vs. 2621), at least for our topics. 
Therefore, while it is encouraging that LLMs have become more epistemically diverse over time, there is still a gap, which may risk future epistemic problems as LLMs come to replace other means of knowledge acquisition, like search.\footnote{For a comparative study demonstrating how our methodology offers more informative results than existing diversity measurement approaches, please see \autoref{sec:sem_sim_compare}.}

\subsection{\Divtype{} Diversity Across Settings}

Our next focus is on \textbf{RQ2: What is the impact of generation setting on \divtype{} diversity?} We look at \divtype{} diversity across three settings: parametric knowledge in instruction fine-tuned models (IFT), the same models with retrieval-augmented generation (RAG), and the traditional search engine baselines. From \autoref{fig:id_by_model}, we see that, on our topics, RAG appears to have a strong positive impact on diversity, and search tends to be better than both IFT and RAG. To test if this is statistically significant, we use a linear mixed effects regression model with HSD as the dependent variable, generation setting as a categorical fixed effect, and the model as a random effect in order to control for the baseline diversity of each model. 

\begin{table}[t]
\centering
\setlength{\tabcolsep}{2pt}
\rowcolors{2}{gray!15}{white}
\begin{tabular}{@{}lll@{}}
\toprule
Predictor  & $\widehat{\beta}$ & $p$-value         \\ \midrule
\cellcolor{black!75}\color{white}Intercept (Med.)  &  910.34($\pm228.07$) & $p\ll1\mathrm{e}{-3}$ \\ \midrule
\cellcolor{black!75}\color{white}Small   & +219.39($\pm44.28$)& $p\ll1\mathrm{e}{-3}$ \\
\cellcolor{black!75}\color{white}Large & -185.93($\pm45.46$)& $p\ll1\mathrm{e}{-3}$          \\
\bottomrule
\end{tabular}

\caption{Estimated coefficients ($\hat{\beta}$) w/ 95\% CIs, and $p$-values of a linear mixed effects model with the model size (Small, Medium, and Large) and setting (IFT and RAG) as fixed effects and the model/version as a random effect. The dependent variable is HSD. We find an inverse relationship between model size and diversity.}
\label{tab:size_regression}
\end{table}

From \autoref{tab:ift_vs_rag}, we see that \textit{RAG and search yield significantly more diverse outputs than parametric memory}, at least for our topics. This highlights the potential of RAG in ensuring that models are epistemically diverse; however, this may depend on RAG databases remaining human written. If traditional search platforms and RAG knowledge bases become dominated by LLM-generated content, the diversity benefits of RAG could be erased, risking knowledge collapse. Therefore, given the benefits that RAG can endow to \divtype{} diversity, we recommend RAG databases remain diverse and limited in the quantity of LLM-generated text included. 

\subsection{\Divtype{} Diversity Across Models}

We now ask \textbf{RQ3: How does model selection impact \divtype{} diversity?} %
From \autoref{fig:id_by_model}, we see that model size appears to have a  \textit{negative} impact on diversity. To test this, we use a linear mixed effects model with the HSD as the dependent variable, the model size binned into three categories (Small, Medium, and Large; see \autoref{tab:model_card}) and the generation setting (IFT or RAG) as fixed effects, and the model version as random effects. The results in \autoref{tab:size_regression} show that \textit{model size has a statistically significant negative effect on epistemic diversity}, echoing contemporary work showing that large models memorize more~\cite{DBLP:journals/corr/abs-2505-24832} and can be less diverse~\cite{zhang2025noveltybench,DBLP:journals/corr/abs-2410-22592} than small models. We hypothesize that this occurs due to their greater capacity and thus greater propensity to memorize~\cite{DBLP:journals/corr/abs-2505-24832}, narrowing the output token distribution given prompts about the same topic. This has practical implications: the performance gains from large models may come with a cost in the form of decreased diversity in the information those models can generate.

\subsection{\Divtype{} Diversity Across Countries}
\label{sec:culture_results}

\begin{figure}[t]
        \centering

        \includegraphics[width=0.98\linewidth]{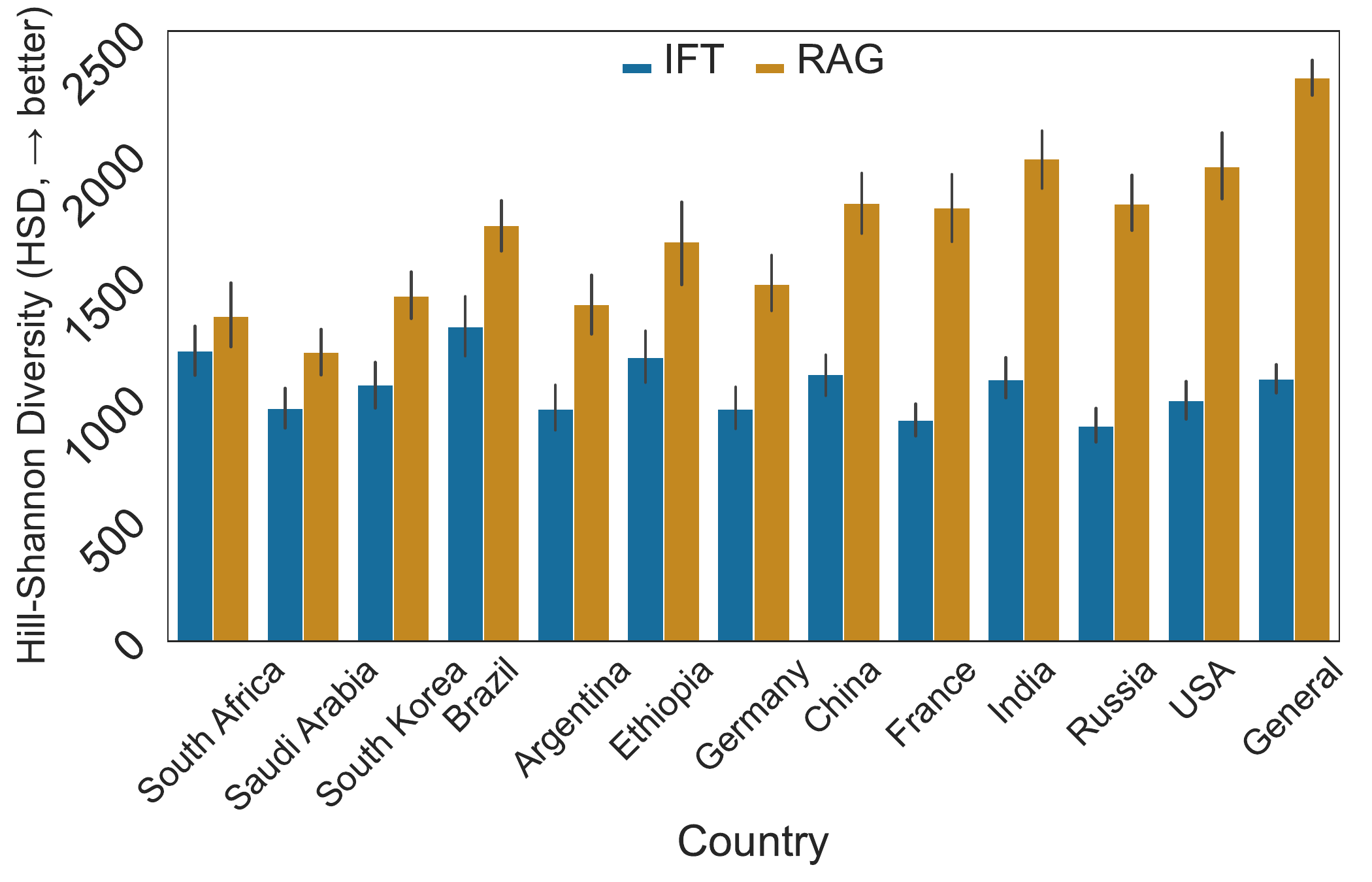}

        \caption{Average diversity per country with bootstrapped 95\% confidence intervals. Bars sorted by the difference between RAG and IFT diversity. RAG appears to have an uneven impact on different countries: the US and general topics see higher benefit.}
        \label{fig:country_heatmap}
\end{figure}

\noindent Aggregate diversity scores can hide inequality in knowledge representation across countries. Having established that diversity is improving in aggregate but lags search, we now ask a harder question: \textbf{RQ4: Whose knowledge is represented by LLMs?} We examine this in two ways: how the country associated with a topic affects diversity (RQ4.1), and whether LLMs reflect English or local-language knowledge for country-specific topics (RQ4.2).

First, \textbf{RQ4.1: how does the source country of a topic impact \divtype{} diversity?} 
To answer, we plot HSD vs. country as a bar chart in \autoref{fig:country_heatmap}. The country of each topic is selected if either (1) a historical event occurred in a particular country or (2) a public figure is primarily associated with a certain country. We plot separate bar charts for the IFT and RAG settings. For IFT only, the country does not have a strong impact on HSD, as most countries are within the same 95\% confidence intervals, with the exception of Ethiopia, Saudi Arabia, and Brazil, which are higher. However, each country is impacted differently by RAG. In particular, the USA, India, Russia, France, and China, as well as general concepts, see more benefit from RAG. This is likely because the search we use to acquire documents for RAG is performed in the US region, which may under-represent topic-specific information from certain countries.
To check this, we measure Pearson's $r$ between the \divtype{} diversity of search and the difference between RAG diversity and IFT diversity, averaging by country. 
We find that these two are strongly correlated at 0.73 ($p$ < 0.01, N=13), meaning that \textit{when the diversity of the RAG source for a particular country's set of topics increases, the improvement in diversity from IFT to RAG is likewise greater.} 

\begin{figure}[t]
        \centering
        \includegraphics[width=0.98\linewidth]{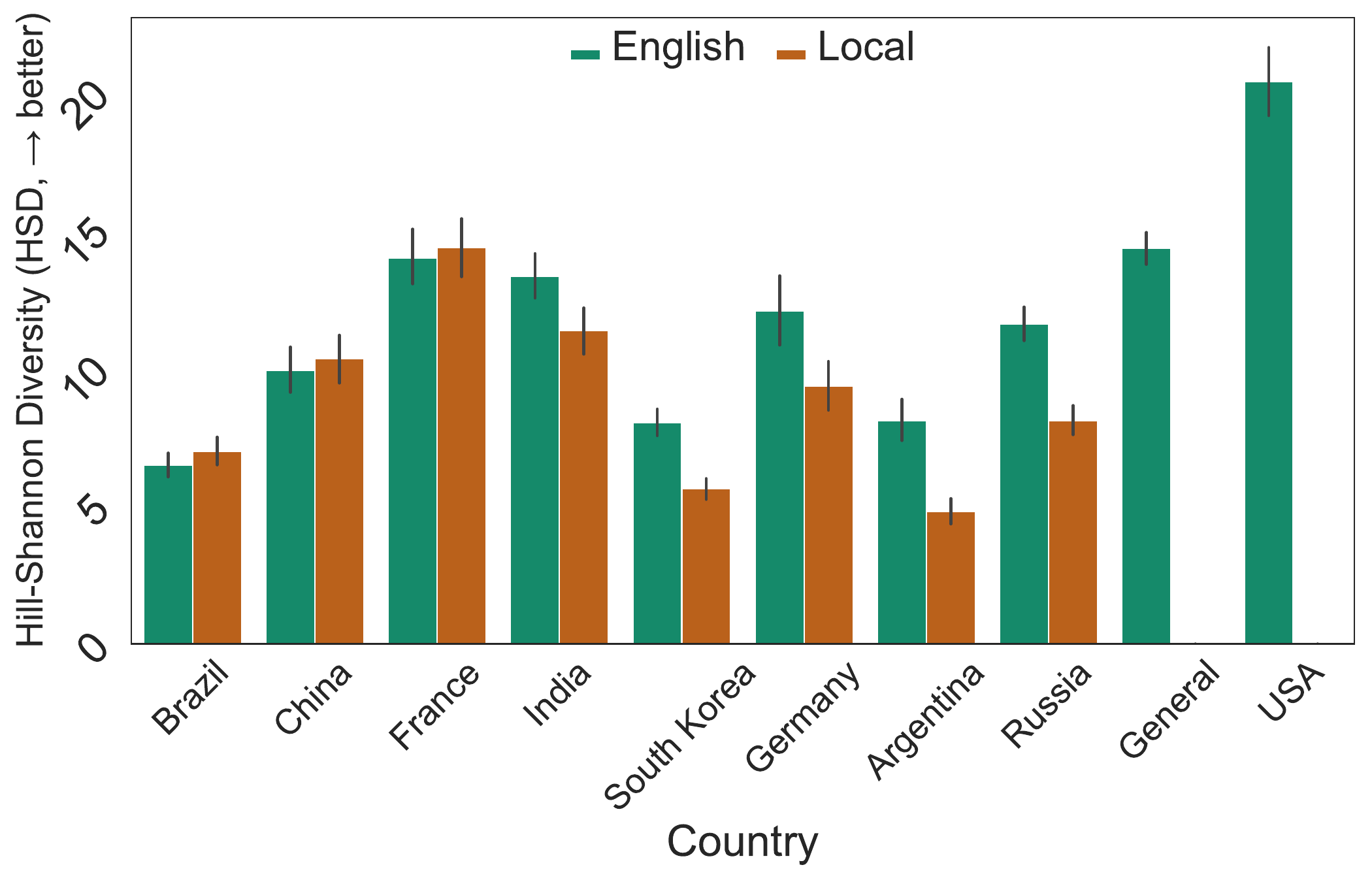}
        \caption{Comparison of the diversity of claims generated by LLMs matched to English and local language Wikipedia. To capture minimal representativeness \cite{DBLP:journals/ais/Peterson25}, we restrict the measurement of HSD only to matched claims. Bars are sorted by the difference between English and local language HSD.}
        \label{fig:id_by_country}
\end{figure}

Second, \textbf{RQ4.2: to what extent is English vs. local language knowledge represented?} 
To answer this, we match LLM claims to Wikipedia claims in both English and local languages using InfoGap, which uses GPT-4 and multilingual sentence embeddings to match claims between Wikipedia pages in different languages (\citealt{DBLP:conf/emnlp/SamirPFST24}, see \autoref{sec:infogap} for details). We use 8 countries in our study where the local language Wikipedias are in the top 20 in terms of active users, with the exception of Saudi Arabia for which the local language pages had minimal content. We use only the topics which are relevant for these countries, and compare to the USA and general topics as baselines.
In order to quantify the extent to which Wikipedia claims appear in model outputs, we measure the HSD of each model after filtering out claims that are not matched to either the English or local language Wikipedia. This is equivalent to ``minimal representativeness'' outlined by \citet{DBLP:journals/ais/Peterson25}, which defines that, for a set of items $i$ that are relevant to a particular task 
``any item should have at least \textit{some} chance of appearing in the LLM output.''
 Results are plotted as a bar chart in \autoref{fig:id_by_country}, aggregated across models and topics for each country.

We observe a \textit{statistically significant gap between English and local language representation for 5 out of 8 countries.} In no case is local representation statistically significantly greater than English representation. This means: a person asking an English LLM about one of our topics, in English, is more likely to receive a response grounded in how that topic is described in English than in how it is understood and documented locally. Additionally, representation for the USA specific topics is statistically significantly greater than that for every other country. This suggests that current English LLMs fail to present knowledge which has not been presented in English for certain countries, echoing recent work on cultural bias in LLM encoded knowledge~\cite{DBLP:conf/emnlp/WangPLB25} and highlighting a risk in terms of knowledge \textit{erasure}, and thus collapse.

\section{Discussion and Conclusion}

This paper asks: how diverse are the claims presented in LLM outputs, over time and across countries?
We answer this by proposing a new methodology for measuring \divtype{} diversity and studying \textit{what} claims are generated and \textit{whose} knowledge is represented in terms of country across 27 LLMs and 155 topics, analyzing 1.7M responses and 70M claims. Our results reveal a nuanced trajectory which the prevailing work on diversity and knowledge collapse has not found~\cite{DBLP:journals/corr/abs-2510-22954, DBLP:journals/ais/Peterson25}: \divtype{} diversity has been improving in the past few years, but there remain salient gaps. All models are still less diverse than a traditional web search on our topics. For users, this means that relying on LLMs as a primary knowledge source means accepting less diverse information than a simple web search could reveal. For model developers, scaling alone does not improve epistemic diversity, and in fact can harm it, while RAG with diversified sources can provide a robust way to increase diversity. The choice of RAG database is also critical for improving \divtype{} representation across countries, as there is a strong correlation between RAG source diversity and downstream diversity. 
Finally, compared to Wikipedia in both English and local languages for country-specific topics, the claims generated in our study reflect English language claims more than local language claims. This highlights a need to shift focus from measuring and improving diversity in the abstract to measuring and improving diversity for the benefit of the breadth of different communities that use LLMs in practice.

\section*{Acknowledgements}
DW was supported by a Danish Data Science Academy postdoctoral fellowship (grant: 2023-1425). JM was supported by the Stanford Interdisciplinary Graduate Fellowship, the Stanford Center for Affective Science Graduate Fellowship, and the Future of Life Institute Vitalik Buterin PhD Fellowship. SM, MA and SY were supported by the Pioneer Centre for AI, DNRF grant number P1. 

\section*{Limitations}
In terms of topical coverage, we use a combination of randomly selected and hand curated topics. We attempted to select topics that broadly cover general concepts, historical events, and important figures across 12 countries, including both controversial and non-controversial topics. However, manual selection of topics may impact the generalizability of our results. Additionally, our notion of ``whose'' knowledge is coarse grained, focusing on a country-level signal, and for some countries in the study selection of topics is performed from an outsider perspective. An improvement to our setup could be to cover more topics by identifying what people typically search for using LLMs and randomly sampling from that pool of queries. Other notions of popularity can also be examined in terms of Wikipedia page views or trending search keywords. There is also a tradeoff in our selection of topics based on a Wikipedia content rating of ``C'' in that we exclude very long-tail and niche topics from our study, which could be informative in their own right. We also use only 200 different prompt templates due to computational constraints. Greater coverage could be achieved through more sampling.

Regarding our decomposition and clustering algorithms, while we find through manual and automated evaluation that though the quality tends to be high, performance is not perfect. Therefore, there is some irreducible noise in the results. Additionally, we adopt a strict notion of semantic equivalence for clustering in our methodology, while many real world claims may have some partial overlap in information which could be useful to capture when measuring diversity. Stronger NLI approaches also exist, such as using an LLM, but we choose a validated and more efficient BERT based approach at the cost of some accuracy. To illustrate the cost, we perform $\approx$744M NLI inferences in our study; assuming $\approx$40 tokens per inference, with GPT-5, this would cost $\approx$\$44,640.

Next, we choose a RAG setup which is intended to simulate a real-world RAG scenario, while actual RAG implementations will inevitably vary. It would therefore be useful to examine the utility of RAG in a more systematic way, benchmarking several implementations. Finally, while \divtype{} diversity is important, it should also be contextualized with other important aspects of knowledge such as factuality and relevance. %

\bibliography{acl_latex}

\appendix

\section{Full Evaluation Details}
\label{sec:annotation_instructions}

\subsection{Decomposition}
To evaluate the quality of decomposition, we develop three initial decomposition prompts (P1-P3) and have two independent annotators label (1) the \textit{quality} of individual decomposed claims on a Likert scale from 1-5, and (2) how many claims in the original input chunk are \textit{missing} from the list of decomposed claims. On our scale, 1 means that the decomposed claim is not inferable from the original text, and 5 means that the claim is fully inferable and decontextualized. We label 100 input chunks for quality (1) and 100 instances for missing (2), all anonymized for the prompt type (P1-P3). Between the two annotators, we achieve a Pearson correlation of 0.74 for (1) and 0.58 for (2), indicating moderate to strong agreement and the overall feasibility of the task. In order to resolve disagreements, we have a third annotator label those instances where the original annotators disagreed, followed by rounds of discussion and feedback among the annotators to decide on the final labels. 

We then set up an LLM-as-a-judge using G-Eval~\cite{DBLP:conf/emnlp/LiuIXWXZ23}, which achieves 0.6 Pearson correlation with the human labels for decomposition quality and 0.68 for the number of missing claims.
We use this to automatically label 6k instances across three decomposition prompt variants (P1-P3). The results are given in \autoref{tab:decomposition_eval}. We use P3 for the final decomposition in order to prioritize quality.

\paragraph{Decomposition Quality Instructions} You will be given a short piece of text (around 3 sentences) about a provided topic, and a list of atomic claims. Your task is to annotate to what degree each claim is represented in the original short piece of text. Each claim will be rated on a scale from 0 to 5, where each value has the following meaning:
\begin{itemize}[noitemsep]
    \item 0 - EMPTY (skip)
    \item 1 - The claim is totally irrelevant to the original piece of text OR does not explicitly talk about the provided TOPIC
    \item 2 - The claim is incomplete or somewhat included in the original piece of text and missing clearly important context
    \item 3 - The claim is included in the original piece of text but missing some potentially important context. This includes claims which could be inferred from the original context but aren't explicitly stated (e.g., "The lifecycle of plastic includes production." being inferred from "It addresses the entire lifecycle of plastic, from production and consumption to disposal and recycling.")
    \item 4 - The claim is included in the original piece of text and is missing only unimportant context (the most important information is represented)
    \item 5 - The claim is included in the original piece of text and no context is missing
\end{itemize}

\paragraph{Decomposition Missing Claims Instructions}
In addition, after each group of claims you will be asked to enumerate any missing claims which are relevant to the topic. Please compare the text to the claims, and identify any relevant claims that were not listed. You just need to provide the NUMBER of claims that you estimate are missing (you do not need to write the claims down, just the number). 

\subsection{Clustering}
Two independent annotators label: (1) the degree of \textit{cohesion} among the clustered claims (Likert score between 1 and 5), and (2) whether any claims are \textit{missing} from a given cluster (binary yes/no).
We define cohesion as the degree to which each piece of text in the cluster conveys at least one piece of information in common.
We define missing whether semantically similar claims which were not predicted to be included in a given cluster (i.e., singletons) \textit{should have} been grouped with said cluster.
We label 100 instances of cohesion (1) and 360 instances of missing (2), achieving a Pearson correlation of 0.44 for (1) and 87\% agreement accuracy for (2). Due to the moderate correlation in (1), we had a follow-up discussion and tie-breaking session which included a third annotator. Using the resolved annotations we then set up an LLM-as-a-judge using G-Eval in order to annotate and evaluate the rate of cluster cohesion and missing sentences in our final clusters. We develop prompts that achieve 0.68 Pearson $r$ with the human labels for measuring cluster cohesion and 83.5 weighted F1 score for measuring binary missing claims. 15,000 samples of 3-5 sentences are then automatically labeled from clusters acquired using \autoref{alg:clustering}, where semantic similarity for task (2) is measured using S-BERT. We acquire a cluster cohesion score of \textbf{4.17} and a missing rate of \textbf{13.79\%}, indicating that the cluster cohesion is generally of high quality while missing only a relatively small proportion of singletons. We compare this to a semantic similarity only approach using DBSCAN in Appendix \autoref{sec:sem_sim_compare_tropes}~\cite{DBLP:conf/emnlp/0001ABYBA24}, showing that cluster cohesion degrades to 3.48.

\paragraph{Clustering Cohesion Instructions}
You will be shown groups of text. Your job is to rate each group of text for the degree to which each piece of text in the group conveys at least one piece of information in common. Each group will have a maximum of 10 pieces of text and a minimum of 3 pieces of text. Rate each group as follows: 
\begin{itemize}[noitemsep]
    \item 1 - None of the sentences have any information in common
    \item 2 - Some of the sentences have information in common but most sentences convey something different
    \item 3 - About half of the sentences convey one thing in common
    \item 4 - Most of the sentences convey at least one piece of information in common
    \item 5 - All of the sentences convey at least one piece of information in common
\end{itemize}

\paragraph{Clustering Missing Sentences Instructions}
You will then be shown 10 additional pieces of text. Your task will be to determine whether or not these pieces of text belong in the group. You will rate each piece of text as follows:
\begin{itemize}[noitemsep]
    \item 0 - The sentence does not convey any information in common with the group
    \item 1 - The sentence conveys the same common information shared by the group; the sentence may convey some more specific or extra information, but the core information of the group is conveyed by the sentence as well
\end{itemize}

\section{Prompts}
\label{sec:prompts}
Here we show decomposition prompts P1 (\autoref{fig:decomposition_prompt_p1}) and P2 (\autoref{fig:decomposition_prompt_p2}) used in our evaluation. Decomposition prompt P3, which we use in the final experiments, is the same as P2 but includes three in-context examples. All three prompts can be found in our linked code repository.\footnote{\codeurl{}}

\begin{figure}[h]
    \centering
    \begin{sharp_box}
    Please breakdown the following paragraph into a list of independent facts. Please return the facts as a list. Please do not number the list. Please simply separate each fact by a new line character.
    
    \texttt{\{content\}}
    \end{sharp_box}
    \caption{Decomposition prompt P1.}
    \label{fig:decomposition_prompt_p1}
\end{figure}

\begin{figure}[h]
    \centering
    \begin{sharp_box}
    I am going to show you a piece of TEXT that talks about the following ISSUE: \texttt{\{issue\}}.
Please read every single sentence in the TEXT and then decompose each sentence into individual, atomic STATEMENTS.
Please make the STATEMENTS totally faithful to the original TEXT.
Please only include STATEMENTS about {issue}. If there are no sentences in the TEXT which include statements about {issue}, please just say EMPTY.
Please ignore sentences which talk about the TEXT itself, for example "In sum, we have talk about \texttt{\{issue\}}."
Please only include STATEMENTS which are fully supported by the TEXT.
Please don't hallucinate information.
Please do not repeat STATEMENTS.
Please return the STATEMENTS as a list. Please do not number the list. Please simply separate each STATEMENT by a new line character.
Please double check that each of the STATEMENTS in the list is entailed and factually accurate with the text.
If the TEXT is presented as a list of bullet points, please make sure to decontextualize the bullet points (don't just list them).
Here is the text:

\texttt{\{content\}}
    \end{sharp_box}
    \caption{Decomposition prompt P2.}
    \label{fig:decomposition_prompt_p2}
\end{figure}

\section{Clustering Algorithm}
\label{sec:algorithm}
\autoref{alg:clustering} outlines our approach for \textbf{clustering} the decomposed claims $\Claims_{\Model{}_{\Topic}}$ into semantically equivalent meaning classes $\Espace{\Model{}_{\Topic}}$.
\begin{algorithm}[t]
\caption{Clustering algorithm}
\small
\label{alg:clustering}
\SetKwInOut{Input}{input}
\SetArgSty{textnormal}
\DontPrintSemicolon
\Input{Decomposed corpus $\Claims{}$ with claims $c_{j}$; Max retrieval length $N$}
\CommentSty{\# Previous claims}

$A \gets [C[0]]$;

\CommentSty{\# Claim cluster}

$L \gets [0]$;

\ForAll{$c_{j} \in \Claims[1:]$}{
    \CommentSty{\# Sort by cos similarity}
    
    candidates $\gets$ most\_similar($c_{j}$, $A$)[:$N$];

    scores = [];
    
    \ForAll{$c_{k} \in$ candidates}{
        \If{entails($c_{k}$, $c_{j}$) \& entails($c_{j}$, $c_{k}$)}{
            scores.append(entail\_prob($c_{k}$, $c_{j}$) * entail\_prob($c_{j}$, $c_{k}$));
        }
        
    }

    \If{len(scores) == 0} {

        $i \gets \max(L) + 1$;
    }
    \Else{
        
        $i \gets$ cluster($\max$(scores));
    }

    $L$.append($i$);
    
    $A$.append($c_{j}$);
    
}

\CommentSty{\# Count the number of elements in each unique cluster}

$\Espace{} \gets \text{group}(L)$;

\CommentSty{\# Break up large clusters}

\ForAll{$x \in \Espace{}$}{
    \If{len($x$) > 100}{
        $\Espace{}'$ = DBSCAN($x$);
        
        $\Espace{}$.pop($x$);
        
        $\Espace{}$.extend($\Espace{}'$);
    }
}

\Return{} $\Espace{}$;

\end{algorithm}

\section{Topics}
\label{sec:topics}
\textbf{USA} Donald Trump;
Claude Shannon;
Barack Obama;
Frederick Douglass;
Bob Dylan;
Iran hostage crisis;
9/11;
American civil war;
Roe v. Wade;
Independence Day (United States);
\textbf{Argentina} Javier Milei;
Juan Domingo Perón;
Cristina Fernández de Kirchner;
Jorge Rafael Videla;
Diego Maradona;
Falklands War;
Infamous Decade (Argentina);
May Revolution;
1998–2002 Argentine great depression;
Cadet scandal in Argentina;
\textbf{Ethiopia} Kenenisa Bekele;
Sophia Bekele;
Meles Zenawi;
Mengistu Haile Mariam;
Kebede Michael;
History of Ethiopia;
Ethiopian Civil War;
Tigray War;
History of the Federal Democratic Republic of Ethiopia;
COVID-19 pandemic in Ethiopia;
\textbf{Korea} Syngman Rhee;
Yuna Kim;
Kim Soo-hyun;
Sejong the Great;
Ban Ki-moon;
Gwangju uprising;
Seollal;
South Korean March First Movement;
K-pop;
Sinking of MV Sewol;
\textbf{Russia} Vladimir Putin;
Joseph Stalin;
Fyodor Dostoevsky;
Alexei Navalny;
Garry Kasparov;
October Revolution;
Soviet Invasion of Poland;
Sputnik 1;
1993 Russian constitutional crisis;
Wagner Group rebellion;
\textbf{France} Charles de Gaulle;
Marine le Pen;
Albert Camus;
Michel Foucault;
Napoleon;
Dreyfus affair;
the French Revolution;
Charlie Hebdo shooting;
Arenc affair;
Yellow vests protests in France;
\textbf{India} Jallianwala Bagh massacre;
Partition of India;
2002 Gujarat riots;
Indian Rebellion of 1857;
Narendra Modi;
Jalal-ud-din Muhammad Akbar;
B. R. Ambedkar;
Indira Gandhi;
Vallabhbhai Patel;
Non-cooperation movement (1919–1922);
2008 Mumbai attacks;
\textbf{Saudi Arabia} Assassination of Jamal Khashoggi;
Al-Yamamah arms deal;
Khobar Towers bombing;
Proclamation of the Kingdom of Saudi Arabia;
Riyadh International Book Fair;
Ibn Saud;
Hatoon al-Fassi;
Manal al-Sharif;
Ayatollah Sheikh Nimr Baqir al-Nimr;
\textbf{South Africa} 2021 South African unrest;
1999 Tempe military base shooting;
Soweto uprising;
Miriam Makeba;
Nadine Gordimer;
Nelson Mandela;
Evelyn Mase;
Christiaan Barnard;
Crizelda Brits;
Rand Rebellion;
South African Border War
\textbf{Brazil} Paraguayan War;
1937 Brazilian coup d'état;
Revolution of the Ganhadores;
Brazilian Carnival;
Dilma Rousseff;
Pedro II of Brazil;
José Paranhos, Viscount of Rio Branco;
Indigenous peoples in Brazil;
Carmen Miranda;
Carlos Chagas;
Mensalão scandal;
\textbf{China} 1989 Tiananmen Square protests and massacre;
2019–2020 Hong Kong protests;
Annexation of Tibet;
Qingming Festival;
2010 Yushu earthquake;
Jinan incident;
Du Fu;
Xi Jinping;
Chinese Communist Party;
Ai Weiwei;
Soong Ching-ling;
\textbf{Germany} Berlin Wall;
Nuremberg trials;
Night of the Long Knives;
Clara Josephine Schumann;
Wilhelm Richard Wagner;
Frederick the Great;
Karl Marx;
Eschede train disaster;
Frauke Petry;
Ernst Nolte;
\textbf{General} nuclear weapons;
slavery;
pornography;
marriage;
white supremacy;
international relations;
prisons;
domestic violence;
patriotism;
same-sex marriage;
free speech;
political corruption;
universal basic income;
global hunger;
plastic waste;
political correctness;
fascism;
racism;
colonialism;
the impact of climate change;
democracy;
feminism;
human rights;
genocide;
war;
censorship;
artificial intelligence;
renewable energy;
capitalism;
autonomous vehicles;
misinformation;
affirmative action;

\begin{table*}[!t]
\centering
\begin{tabular}{ccllll}
\toprule
\textbf{Family} & \textbf{Ver.} & \textbf{Size} & \textbf{Release Date} & \textbf{Endpoint} \\
\midrule
\multirow{9}{*}{Qwen} 
& 1.5     & 7B (S)   &  February 2024       & Qwen/Qwen1.5-7B-Chat\\
& 1.5     & 14B (M)  &         & Qwen/Qwen1.5-14B-Chat\\
& 1.5     & 72B (L) &         & Qwen/Qwen1.5-72B-Chat \\ \cdashline{2-5}
& 2.5     & 7B (S)  & September 2024 & Qwen/Qwen2.5-7B-Instruct\\
& 2.5     & 14B (M) &         & Qwen/Qwen2.5-14B-Instruct\\
& 2.5     & 72B (L)  &         & Qwen/Qwen2.5-72B-Instruct\\\cdashline{2-5}
& 3       & 8B (S)   & April 2025 & Qwen/Qwen3-8B\\
& 3       & 14B (M)  &         & Qwen/Qwen3-14B\\
& 3       & 32B (L)  &         & Qwen/Qwen3-32B\\ \midrule
\multirow{8}{*}{Gemma} & 1       & 2B (S)   & February 2024        & google/gemma-2b-it \\\cdashline{2-5}
& 1.1     & 7B (M)   &  April 2024       & google/gemma-1.1-7b-it \\\cdashline{2-5}
& 2       & 2B (S)  & July 2024        & google/gemma-2-2b-it \\
& 2       & 9B (M) &  June 2024       & google/gemma-2-9b-it \\
& 2       & 27B (L) &         & google/gemma-2-27b-it \\\cdashline{2-5}
& 3       & 1B (S)  & March 2025        & google/gemma-3-1b-it \\
& 3       & 12B (M) &         & google/gemma-3-12b-it \\
& 3       & 27B (L) &         & google/gemma-3-27b-it \\\midrule
\multirow{9}{*}{Llama} & 2       & 7B (S)   & July 2023        & meta-llama/Llama-2-7b-chat-hf \\
& 2       & 13B (M)  &         & meta-llama/Llama-2-13b-chat-hf \\
& 2       & 70B (L)  &         & meta-llama/Llama-2-70b-chat-hf \\\cdashline{2-5}
& 3.1     & 8B (S)   & July 2024 & meta-llama/Llama-3.1-8B-Instruct \\
& 3.1     & 70B (L)  &         & meta-llama/Llama-3.1-70B-Instruct \\\cdashline{2-5}
& 3.2     & 3B (S)   &  September 2024       & meta-llama/Llama-3.2-3B-Instruct \\\cdashline{2-5}
& 3.3     & 70B (L) & December 2024 & meta-llama/Llama-3.3-70B-Instruct \\
\midrule
\multirow{3}{*}{OpenAI**} & 3.5    & Undisclosed   & November 2022  & gpt-3.5-turbo-0125\\\cdashline{2-5}
 & 4   & Undisclosed & August 2024 & gpt-4o-2024-08-06\\\cdashline{2-5}
 &  5   & Undisclosed & August 2025 & gpt-5-2025-08-07\\
\bottomrule
\end{tabular}
\caption{List of different model versions (Ver.), sizes, year of release, along with model endpoints used either from HuggingFace or the respective API endpoint. ** is for closed-weight models. In terms of model size: Small $\le8B$, $9\le$Medium$<27B$ and Large$\ge27B$. As Gemma models tended to be smaller we include one 7B model (Gemma 1.1 7B) in the Medium category for \autoref{fig:id_by_model}.}
\label{tab:model_card}
\end{table*}

\section{Comparison to Semantic Similarity Only}
\label{sec:sem_sim_compare}

Our aim, reflected in our problem setup and formulation, is to measure how diverse are the unique pieces of knowledge that different people can be exposed to when interacting with an LLM. To contrast our methodology with similar contemporary work on diversity~\cite{DBLP:journals/corr/abs-2503-20062,DBLP:conf/emnlp/0001ABYBA24,DBLP:journals/corr/abs-2510-22954}, an alternative version of our setup would be to partition claims based solely on semantic similarity, i.e., the distance between claims in a semantic embedding space. However, semantic similarity can only describe the degree to which two pieces of text are functionally similar, and only in extreme cases whether they convey the same piece of information.
To illustrate: the phrases ``Claude Shannon is the father of information theory'' and ``Claude Shannon is not the father of information theory'' have high semantic similarity (0.94)\footnote{Huggingface ID: \texttt{all-MiniLM-L6-v2}}, even though they present opposing claims. Similarly, ``Claude Shannon is the father of quantum theory'' is an entirely different claim from ``Claude Shannon is the father of information theory,'' but these two also have high semantic similarity (0.805). Therefore, while semantic similarity can be useful in order to find candidate pairs which are semantically equivalent, partitioning based on similarity alone would lead to classes which contain potentially many different pieces of knowledge, thus underestimating diversity and not reflecting our definition of meaning classes.%
To highlight this in a concrete way,  we now contrast our methodology with two recent popular approaches to diversity measurement which use semantic similarity, in order to demonstrate what is gained by focusing on semantic equivalence of claims as the unit of knowledge.

\subsection{LLM Tropes~\cite{DBLP:conf/emnlp/0001ABYBA24}} 
\label{sec:sem_sim_compare_tropes}
What about partitioning into classes based on semantic similarity? We now compare to the approach from \citet{DBLP:conf/emnlp/0001ABYBA24} which does exactly this. This is performed by taking the decomposed claims from our study, transforming them into sentence embeddings, and using DBSCAN~\cite{DBLP:conf/kdd/EsterKSX96} to partition claims into unique classes. Note that DBSCAN parameters require tuning --- for the sake of comparison, we selected reasonable parameters which resulted in a comparable number of clusters to the semantic equivalence case (eps of 0.2, min size of 2), but better results could potentially be achieved by hyperparameter tuning. After partitioning claims as such, we perform the same LLM-as-a-judge evaluation as in \autoref{sec:clustering}, acquiring a cluster cohesion score of 3.48 (vs. 4.17 for our algorithm) and missing singleton rate of 0.4\% (vs. 13.8\% for our algorithm). As such, the recall of these clusters is higher (reflected in fewer singletons which should have been clustered), but the overall cohesion goes down, reflecting a tradeoff in terms of clustering quality.

\subsection{Artificial Hivemind~\cite{DBLP:journals/corr/abs-2510-22954}} 
\label{sec:sem_sim_compare_hivemind}
\begin{figure*}[!h]
        \centering
        \includegraphics[width=0.98\linewidth]{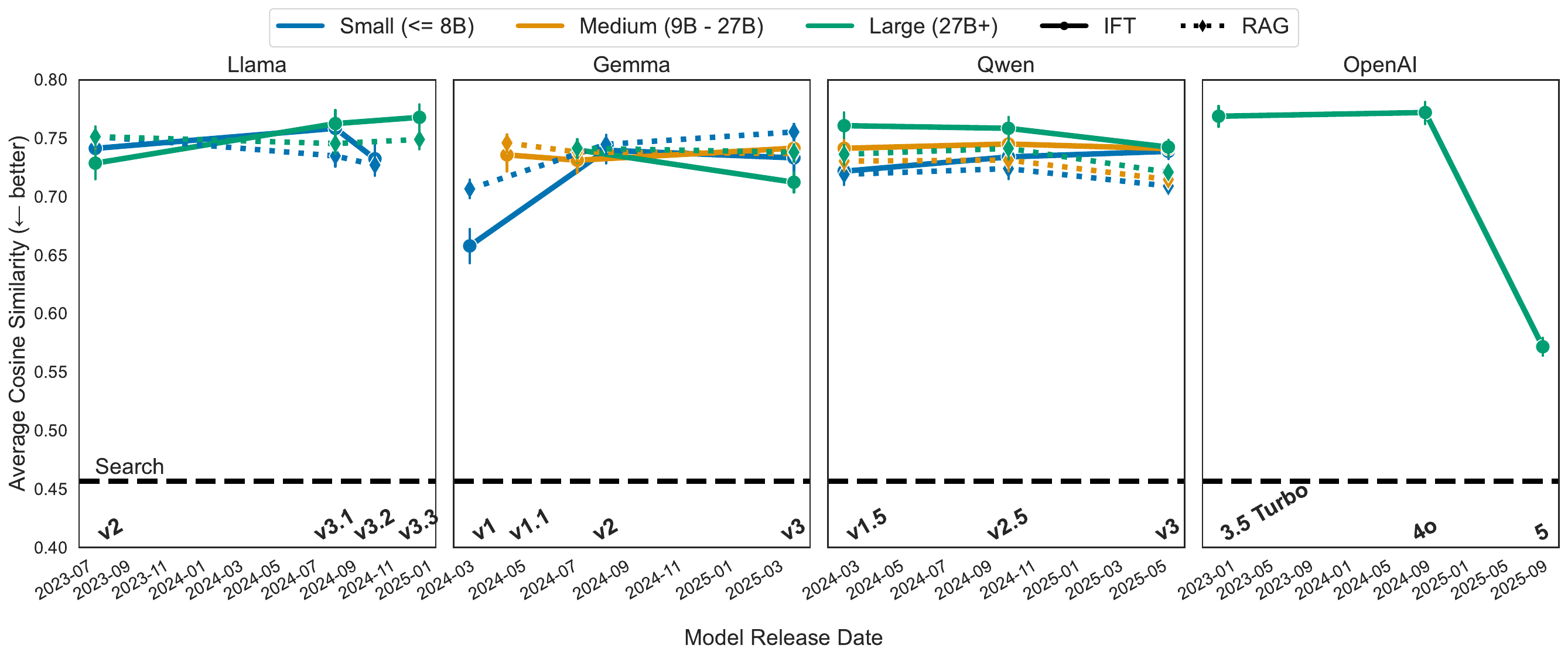}
        \caption{Average pairwise cosine similarity of within-topic model responses
       vs. model release date. Each point is a single model, with lines connecting models of approximately the same size across released versions. Error bars are 95\% boostrapped confidence intervals based on cosine similarity within each topic (N=155). \textbf{Lower scores mean greater diversity}. While we see that search is still much more diverse than LLMs, many of the trends identified via epistemic diversity are lost. 
        }
        \label{fig:cosine_id_by_model}
\end{figure*}

We now compare to the approach used in \citet{DBLP:journals/corr/abs-2510-22954}. In this work, diversity within LLM responses is measured using the average pairwise cosine similarity between responses represented as sentence embeddings. In other words, each LLM response is passed through a sentence embedding model (OpenAI's \texttt{text-embedding-3-small} API in the original paper, Huggingface ID \texttt{: all-MiniLM-L6-v2} here), and then the cosine similarity of each pair of embeddings is measured and averaged. We perform exactly this procedure on the original LLM responses of each model within each topic in our study, and plot diversity over time and across models (exactly as in \autoref{fig:id_by_model}) in \autoref{fig:cosine_id_by_model}.

What does this plot tell us? 
We can make a few claims: models lack semantic diversity to approximately the same degree, there has been little change over time with the exception of Gemma 1 and GPT-5, and traditional search is more semantically diverse than LLMs.
However, in the context of our study on the risk of AI knowledge collapse these results are not particularly meaningful or interpretable. First, compared to our experiment measuring \divtype{} diversity, we lose a lot of valuable information. We see no difference with RAG or by model size, and the general improvement over time that we saw with \divtype{} diversity is lost. This is because semantic embeddings can give us a coarse signal of similarity between two pieces of text when cosine distance is measured, while our methodology isolates a fine-grained and targeted signal, namely, semantically equivalent claims. 
Second, the $y$-axis in \autoref{fig:cosine_id_by_model} is much more difficult to interpret than \autoref{fig:id_by_model} --- whereas when measuring HSD based on claims, we can directly make statements about the effective diversity of claims that different models generate, cosine similarity between sentence embeddings is opaque. For example, the average cosine similarity decreased (i.e., improved) by 0.2 from GPT 4o to GPT-5, but this improvement lacks any practical interpretation other than ``semantic diversity increased to some degree.'' On the other hand, from \autoref{fig:id_by_model} we can say that the effective diversity of claims that one can expect to encounter when interacting with GPT-5 is more than 5$\times$ greater than with GPT 4o for our topics.

\section{Model Details}
\label{sec:model_details}
\autoref{tab:model_card} provides the details of the 4 model families employed in this study, along with demarcation of small, medium, and large sets.

\section{InfoGap Details} 
\label{sec:infogap}
We use GPT-5 to decompose English and local language Wikipedias to individual claims, retrieve candidate matches to LLM generated claims using a multilingual sentence embedding model,\footnote{HuggingFace: sentence-transformers/LaBSE} and predict if the retrieved multilingual claims match to the LLM generated English claims using GPT-5 (over GPT-4 as used in \cite{DBLP:conf/emnlp/SamirPFST24}). We exclude claims generated using RAG to avoid Wikipedia text being included in the generation context.

\section{Generation Details}
We retrieve 40 Google search results for each topic. Over the 155 topics, these 6,200 pages have an average of 12,836 and a median of 2,132 tokens per page. For a fair comparison, we allow each model to generate up to 2,100 tokens for each of the 200 prompt variations. For RAG, we include up to 1,000 additional tokens for context. All generation was run on NVIDIA A100 (40GB) and H100 (80GB) GPUs depending on the size of the model. 70B parameter models required 2 H100s for generation.

\section{Artifacts Statement}
All data artifacts used in this project are open source, publicly available, and intended for research purposes. Generated data and code is available at \codeurl{} under the MIT license.

\section{Use of AI in Research Statement}
We made minimal use of Claude in the final polishing stages of preparing the manuscript. This was primarily in getting feedback on the draft in terms of potential ways to improve the figures, consistency of claims, and framing.

\end{document}